%% file: ms.tex
\documentclass[11pt]{article}
\usepackage{coling2020}
\usepackage{times}
\usepackage{url}
\usepackage{latexsym}
\usepackage{graphicx}
\usepackage[ruled,vlined]{algorithm2e}
\usepackage{comment}
\usepackage{subcaption}
\usepackage{adjustbox}

\colingfinalcopy 

\title{EFSG: Evolutionary Fooling Sentences Generator}

\author{Marco Di Giovanni \\
  Politecnico di Milano / Milan, Italy \\
  {\tt marco.digiovanni@polimi.it} \\\And
  Marco Brambilla \\
  Politecnico di Milano / Milan, Italy \\
  {\tt marco.brambilla@polimi.it} \\}

\date{}

\begin{document}
\maketitle
\begin{abstract}
  Large pre-trained language representation models (LMs) have recently collected a huge number of successes in many NLP tasks. 
  In $2018$ BERT, and later its successors (e.g. RoBERTa), obtained state-of-the-art results in classical benchmark tasks, such as GLUE benchmark. 
  After that, works about adversarial attacks have been published to test their generalization proprieties and robustness.
  In this work, we design Evolutionary Fooling Sentences Generator (EFSG), a model- and task-agnostic adversarial attack algorithm built using an evolutionary approach to generate false-positive sentences for binary classification tasks. 
  We successfully apply EFSG to CoLA and MRPC tasks, on BERT and RoBERTa, comparing performances. Results prove the presence of weak spots in state-of-the-art LMs. 
  We finally test adversarial training as a data augmentation defence approach against EFSG, obtaining stronger improved models with no loss of accuracy when tested on the original datasets. 
\end{abstract}

\input{intro}
\input{relwork}
\input{dataset}
\input{LMs}
\input{algo}
\input{exps}

\input{conclusions}

\bibliographystyle{coling}
\bibliography{biblio}

\input{appendix}

\end{document}

%% file: intro.tex
\section{Introduction}

Recently pre-trained language models (LMs) became popular due to their huge successes on many benchmark natural language processing (NLP) tasks~\cite{devlin2018bert,GPT2noauthororeditor,dai2019transformerxl,lample2019crosslingual}. 
These LMs need an expensive pre-training phase, followed by a relatively cheap task specific fine-tuning. This transfer learning mechanism allows us to use the same pre-trained LMs, theoretically encoding general knowledge about languages, for many different and heterogeneous tasks. 
After BERT~\cite{devlin2018bert}, other LMs have been released, designed as improvements or variations of existing models or pre-training procedures \cite{RoBERTA_DBLP:journals/corr/abs-1907-11692,sanh2019distilbert,martin2019camembert,lan2019albert,le2019flaubert,lewis2019bart}.

The huge proliferation of pre-trained LMs was followed by works analyzing their proprieties and characteristics~\cite{DBLP:journals/corr/abs-1907-11932,DBLP:journals/corr/abs-1905-05950,DBLP:journals/corr/abs-1905-10650,DBLP:journals/corr/abs-1906-04341}. 
These studies peak inside the huge models (hundreds of millions of parameters to train, always increasing), lighting a bit those black boxes and aiming to get better intuitions of what happens. 
Interpretability is essential to build better and stronger LMs, increasing their robustness and generalization abilities. 
However, much work has still to be done. 
Our work is another step towards a better understanding of pre-trained LMs by performing adversarial attacks on them. 

Our contribution can be summarized as follows:\\
1: we design, in an evolutionary fashion, a model- and task-agnostic adversarial attack algorithm, EFSG, generating false-positive sentences that fool LMs; \\
2: we successfully test the algorithm on two LMs (BERT and RoBERTa) trained on two binary classification tasks (CoLA and MRPC), and we compare results; \\
3: we experimentally investigate how hyper-parameters affect the performance of EFSG;\\
4: we apply adversarial training as a defence strategy against EFSG, evaluating how robust the model becomes after targeted data augmentation. 

EFSG is inspired by an evolutionary adversarial attack algorithm applied to image data~\cite{Nguyen2014DeepNN}, in particular to fool AlexNet~\cite{NIPS2012_4824} trained on MNIST~\cite{lecun-mnisthandwrittendigit-2010} and ImageNet~\cite{imagenet_cvpr09}. 
Since pre-trained LMs can be almost compared to AlexNet, but for textual data, we investigate if those new LMs based on transformers have the same weaknesses as the ones based on CNN.

The structure of the paper is outlined here: 
section~\ref{sec:rw} contains related works, 
section~\ref{sec:datasets} the datasets selected and section~\ref{sec:lms} the LMs tested. 
The proposed algorithm (EFSG) is described in detail in section~\ref{sec:method} while in section~\ref{sec:exps} we show the experiments and discuss results. 
We conclude in section~\ref{sec:conclusion}.

%% file: relwork.tex
\section{Related work}\label{sec:rw}

After the success of pre-trained language models, the interest about their analysis and applications increased and studies trying to prove or demystify the goodness of those models have been performed.
Interesting results include how steps of the traditional NLP pipeline couple to layers of BERT model, observing quantitatively and qualitatively that the pipeline is dynamically adjusted to disambiguate information~\cite{DBLP:journals/corr/abs-1905-05950};  
an exhaustive analysis of BERT's attentions heads, collecting similar patterns such as delimiter tokens, positional offsets, noticing linguistic notions of syntax and co-reference~\cite{DBLP:journals/corr/abs-1906-04341}; an analysis on relation representations of BERT, designing a training setup called \textit{matching the blanks}, relying solely on entity resolution annotations~\cite{DBLP:journals/corr/abs-1906-03158}. 

Recently, many works about adversarial attacks on text have been published.
\newcite{Kuleshov2018AdversarialEF} formalized the notion of an adversarial sample on a wide range of tasks, and studied how algorithms affect different models.
\newcite{Alzantot_2018} designed black-box population-based optimization algorithms to generate semantically and syntactically similar adversarial examples that fool sentiment analysis and textual entailment models, but adversarial training fails as a defence mechanism.

Other adversarial attacks include TextBugger~\cite{DBLP:journals/corr/abs-1812-05271}, an effective, evasive and efficient attack framework to trick Deep Learning-based Text Understanding (DLTU) systems; TextFooler~\cite{DBLP:journals/corr/abs-1905-10650}, a successfully simple strong baseline to generate natural adversarial text for text classification and textual entailment;  DeepWordBug~\cite{DBLP:journals/corr/abs-1801-04354}, that generates effectively small text perturbations in a black-box setting forcing a deep-learning classifier to misclassify a text input; HotFlip~\cite{DBLP:journals/corr/abs-1712-06751}, a generator white-box adversarial examples to trick a character-level neural classifier, where the authors prove the effectiveness of adversarial training and demonstrate how their algorithm can be adapted to word-level classifiers.
Interesting also how the construction of an adversarial dataset, exploring spurious statistical cues in the dataset, makes state-of-the-art language models achieve random performances~\cite{DBLP:journals/corr/abs-1907-07355}.
\newcite{DBLP:journals/corr/abs-1710-11342} applied GANs to generate adversarial samples that are natural and legible for black-box classifiers for image classification, textual entailment and machine translation.
Finally, Seq2Sick~\cite{cheng2018seq2sick} generates adversarial examples for sequence-to-sequence models using a projected gradient method combined with a group lasso and gradient regularization, and applies it to machine translation and text summarization tasks. 

Recently, an unified Python framework~\cite{Morris2020TextAttack} has been released, for adversarial attacks, data augmentation, and model training in NLP, implementing the most famous adversarial attack recipes on every model implemented in Huggingface GitHub repository \cite{Wolf2019HuggingFacesTS}.

Our work is highly inspired by~\cite{Nguyen2014DeepNN}, where the authors apply fooling algorithms to convolutional neural networks in order to generate images that humans perceive as white noise, but they are predicted by ImageNet as belonging to a specific class with high probability. 

%% file: dataset.tex
\section{Datasets}\label{sec:datasets}

The General Language Understanding Evaluation (GLUE) benchmark (\cite{DBLP:journals/corr/abs-1804-07461}) is a collection of nine English datasets for training, evaluating, and analyzing natural language understanding systems.
The tasks cover a broad range of domains, data quantities, and difficulties, and are either single-sentence or sentence-pair classification tasks.
For the purpose of this work, we selected two among the nine datasets in the GLUE benchmark: the Corpus of Linguistic Acceptability (CoLA) and Microsoft Research Paraphrase Corpus (MRPC). 
CoLA \cite{COLA_DBLP:journals/corr/abs-1805-12471} consists of $10657$ English acceptability judgments drawn from books and journal articles on linguistic theory. Each example is annotated with whether it is a grammatical English sentence.
MRPC \cite{dolan2005automatically} consists of $5800$ pairs of sentences from news sources on the web. Each example is annotated with whether it captures a paraphrase/semantic equivalence relationship. 

%% file: LMs.tex
\section{Pre-trained Language Models}\label{sec:lms}

Our adversarial attack algorithm (EFSG) is model-agnostic, meaning that it can be applied to any black-box sentence classifier, with no constraints on the internal structure. 
The model should take as input one sentence (or a couple of sentences, depending on the task) and output a score between $0$ and $1$ quantifying the likeliness that the input is in one class or the other. 

Since EFSG follows a black-box approach, any model can be attacked.
To test EFSG algorithm, we select two pre-trained language models that have recently obtained state-of-the-art results: BERT~\cite{devlin2018bert} and RoBERTa~\cite{RoBERTA_DBLP:journals/corr/abs-1907-11692}. 
The pre-trained weights of both BERT and RoBERTa, and the instruction on how to fine-tune them, are available on their respective GitHub repositories \footnote{https://github.com/google-research/bert} \footnote{https://github.com/pytorch/fairseq}. 

\paragraph{BERT} (Bidirectional Encoder Representations from Transformers~\cite{devlin2018bert}) is a deep state-of-the-art language representation model based of Transformers~\cite{TRANSFORMERS_DBLP:journals/corr/VaswaniSPUJGKP17} trained in an unsupervised way on $3.3$ billion tokens of English text. 
The model is designed to be fine-tuned to a specific tasks inserting an additional final layer, without substantial task-specific architecture modifications. 
There are many versions publicly available~\cite{turc2019wellread}, but the two most used ones are: BERT base, $12$ layers, $768$ hidden dimensions, $12$ heads per layer, for a total of $110M$ parameters, and BERT large, $24$ layers, $1024$ hidden dimensions, $16$ heads per layer, for a total of $340M$ parameters.

\paragraph{RoBERTa}~\cite{RoBERTA_DBLP:journals/corr/abs-1907-11692} is an improvement of BERT model through an accurate selection of hyper-parameters and pre-training techniques. The result is a state-of-the-art model with the same architecture as BERT, optimized by carefully investigating every design choice made to train the model. The full set of hyper-parameters used to both train and test the model is reported in the original paper. Like BERT, also RoBERTa has been released in different sizes. In this work we use the base version of RoBERTa.

%% file: algo.tex
\section{Algorithm description}\label{sec:method}

Evolutionary Fooling Sentences Generator (EFSG) is a model- and task-agnostic evolutionary generator of false-positive sentences that fool a language model. 

Evolutionary algorithms are inspired by Darwinian evolution, where the individuals face natural selection and reproduce mixing their features. 
The process follows the principle of the survival of the fittest~\cite{spencer1864principles}, since the ones with higher \textit{fitness} are more likely to survive. 

While in nature the definition of the fittest could be more or less intuitive, applied to natural language processing, this idea deserves a detailed explanation.
In EFSG, individuals are sentences. 
New sentences are generated by randomly mutating old ones (\textit{parent} sentences), in a way similar to individuals generated by mutations (and combinations) of their parents. 

The key idea is the computation of the fitness of a sentence using a language model. 
For the sake of comprehension, we explain this concept through an example.
BERT fine-tuned for CoLA task takes as input a sentence $x_i$ and outputs a score $s_i \in [0, 1]$ representing the confidence of the model that the given sentence is grammatically correct or not. 
Using the output of BERT fine-tuned as the fitness score implies that a sentence is more likely to survive the current epoch if BERT is confident that that sentence is grammatically correct.
Thus, the \textit{fittest} sentence is the sentence that obtains a greater score by BERT.
The same idea can be generalized for any LM and task.

The full algorithm is summarized in Algorithm \ref{alg:fooler}. 
Firstly, one sentence is randomly generated. 
The length of sentences $L$, i.e. the number of words in the sentence, is fixed. 
We randomly pick the words from a distribution $D_{task}$ obtained collecting words from the training set. This assures that only words already seen by the model are used to generate sentences, with probability proportional to the frequency that they appeared in the training set. Note that this distribution $D_{task}$ is task dependent, since different tasks have different training sets, thus $D_{CoLA}\neq D_{MRPC}$. Punctuation is removed, thus the generated sentences are just an ordered collection of words. 

The first sentence is mutated $N$ times obtaining a set of $N$ potentially different sentences. 
A \textit{mutation} is defined as the replacement of a random word (picked uniformly from the available $L$ words in the sentence) with a word from the distribution $D_{task}$. Mutations keep the length of the sentence fixed. 

Then, the LM scores the sentences, returning a set of scores $S^0$.
We use the normalized scores $S^0$ to pick two parents of sentences for the next generation of sentences, e.g. sentence $x_1^0$ is selected as $Parent_1$ with probability $s_1^0$ normalized. 

\textit{Crossover} is implemented by randomly selecting words from parent sentences. The $n$-th word of the children have $50\%$ of probability to be the $n$-th word of $Parent_1$ and $50\%$ of probability to be the $n$-th word of $Parent_2$. Crossover can be avoided by setting the option $doCrossover$ as False.
After Crossover, we apply mutations as described above. 

We iterate these steps $P$ epochs, expecting the average score to increase as long as we keep iterating.

When EFSG algorithm is applied to a sentences-pair classification task, such as MRPC, one sentence $\hat{x}$ is fixed and the other is evolutionary generated. The algorithm is unchanged, but the LM takes as inputs the fixed sentence $\hat{x}$ and the generated one ($s_i^p \leftarrow LM(\hat{x}, x_i^p)$).

We are aware that this procedure is not able to explore the full sentences' space, being dependent on the first sentence generated. However, the algorithm is not expected to be exhaustive, but to find at least some fooling samples.

Our assumption is that, since the generation process is based on random mutation and crossover, the generated sentences will be random collections of words, thus false-positives for any task. 
We are aware that grammatically correct sentences could be generated, but we expect it to be a rare event. To confirm our assumption, samples of sentences for different experiments are reported in tables in the appendix. 

\begin{algorithm}[h]
\SetAlgoLined
 $x_0\leftarrow Initialize(L,D_{task})$ \;
 \For {$i=1,...,N$}{
  $x_i^0\leftarrow Mutate(x_0, D_{task})$\;}
  
 \For{$p=1,...,P$}{
  \For {$i=1,...,N$}{
   $s_i^{p-1} \leftarrow LM(x_i^{p-1})$\;}
  $S^{p-1} = Normalize(s_1^{p-1},...,s_N^{p-1})$\;
  \For {$i=1,...,N$}{
    $Parent_1 \sim S^{p-1}$\;
    \If {doCrossover}{
      $Parent_2 \sim S^{p-1}$\;
      $x_i^p \leftarrow Crossover(Parent_1, Parent_2)$\;
      $x_i^p \leftarrow Mutate(x_i^p, D_{task})$}
    \Else{$x_i^p \leftarrow Mutate(Parent_1, D_{task})$}
    }
  }
 \caption{EFSG algorithm}\label{alg:fooler}
\end{algorithm}

%% file: exps.tex
\section{Experiments and results}\label{sec:exps}
In this section we describe the experiments performed applying EFSG on BERT and RoBERTa for CoLA and MRPC tasks, including hyper-parameters analysis, models comparison and adversarial training. 
We report results as averages of five identical experiments. 

We remark that, when EFSG is applied to MRPC task (classification of a pair of sentences ($\hat{x}$, $x_i$)), it generates one sentence $x_i$ fixing the other one $\hat{x}$. Thus, we initially sampled a random sentence from the training set and we used it in every experiment described below, but in section~\ref{sec:id} (where we explore different choices of $\hat{x}$), for consistency reasons.
The selected $\hat{x}$ sentence reported in Appendix B (table~\ref{tab:first_sen}) with $id=0$. 


\subsection{Model fine-tuning}

BERT and RoBERTa, base and large, models are pre-trained models, downloadable on the official repositories, built for transfer learning. Thus, they require to be fine-tunined for a specific task. 
We fine-tune BERT base and large both on CoLA and MRPC tasks, with an hyper-parameter grid search, as suggested in the original paper~\cite{devlin2018bert}. The best hyper-parameters are reported in Appendix A (table~\ref{tab:hype}). 
We fine-tune RoBERTa base using the hyper-parameters suggested in the official repository\footnote{https://github.com/pytorch/fairseq/blob/master/examples/roberta/README.glue.md}.

\begin{figure}[t]
    \centering
    \begin{subfigure}{.28\textwidth}
      \centering
      \includegraphics[width=\linewidth]{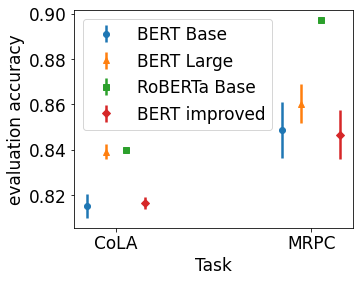}
      \caption{}
      \label{fig:eval_accs}
    \end{subfigure}
    \hfill
    \begin{subfigure}{.28\textwidth}
      \centering
      \includegraphics[width=\linewidth]{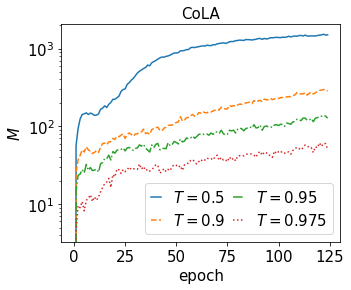}
      \caption{}
      \label{fig:CoLA_threshold_comparison}
    \end{subfigure}
    \begin{subfigure}{.28\textwidth}
      \centering
      \includegraphics[width=\linewidth]{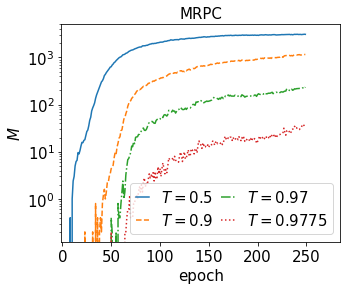}
      \caption{}
      \label{fig:MRPC_threshold_comparison}
    \end{subfigure}
    \caption{Figure (a): Evaluation accuracy of five different fine-tunings of BERT base, BERT large, RoBERTa base and BERT improved (BERT base after Adversarial Training). Figure (b) and (c): average number of fooling sentences $M$ during the fooling algorithm on BERT base for different values of threshold $T_{task}$ (the red dotted line is the threshold ultimately selected).}
\end{figure}

We show evaluation accuracies in figure~\ref{fig:eval_accs} (BERT improved, explained later in section~\ref{sec:at} is fine-tuned using the same hyper-parameters as BERT base in table~\ref{tab:hype} in Appendix A). 
As expected, BERT large accuracy is higher than BERT base accuracy in both tasks, while RoBERTa base is comparable to BERT large for CoLA, while it outperforms it for MRPC.

\subsection{Fooling Threshold}
CoLA and MRPC are both binary classification tasks. 
Pre-trained language models are usually fine-tuned to predict a score $s_i\in[0,1]$ for each sentence $x_i$, for CoLA, or couple of sentences ($\hat{x}$, $x_i$), for MRPC, received as input. 
If the score is higher than a threshold $T_{class}$ usually set to $0.5$, the input is classified as positive, otherwise negative. 
Thus, score (output of the fine-tuned language model) can be thought as the model \textit{confidence} of its prediction. 
The closer the value is to $0$ or $1$, the more confident the model is about its negative or positive prediction respectively. 

Our goal is to generate sentences that are not only classified wrongly, but with high confidence, thus we need to set a different task-dependent threshold value $t_{task}$. 
Since EFSG is designed to generate false positives, i.e. sentences or couples of sentences that are classified positively even if they are not, the threshold value $t_{task}$ should be greater than $0.5$. To set a precise numerical value, we use the scores obtained by sentences in the training set. More precisely, we set $T_{task}$ such that $75\%$ of positive sentences from the training set used to fine-tune the models get a value higher than $T_{task}$. When BERT base is fine-tuned on CoLA, $T_{CoLA}=0.975$, while when BERT base is fine-tuned on MRPC, $T_{MRPC}=0.9775$. When we test BERT large or RoBERTa base, the values obtained are similar so we fix them regardless of the model.

After setting $T_{task}$, we can define as $M$ the number of fooling sentences: generated sentences obtaining a score higher than the selected threshold ($T_{CoLA}$ or $T_{MRPC}$) when evaluated by a fine-tuned LM. We also define $F=\frac{M}{N}$, where $N$ is the number of generated sentences each epoch.

Obviously, the higher the threshold $T_{task}$ the harder it is to generate sentences with a score greater than it. 
In figure~\ref{fig:CoLA_threshold_comparison} and~\ref{fig:MRPC_threshold_comparison} we show magnitudes of $M$ for different thresholds when BERT base is fooled by EFSG on CoLA ($N=5000$, $L=9$) and MRPC ($N=5000$, $L=12$) tasks respectively. Blue solid lines indicate how easy is to generate sentences classified wrongly, while red dotted lines highlight values of $M$ when we set the final thresholds ($T_{CoLA}$ and $T_{MRPC}$).

\begin{figure}[t]
    \centering
    \begin{subfigure}{.24\textwidth}
      \centering
      \includegraphics[width=\linewidth]{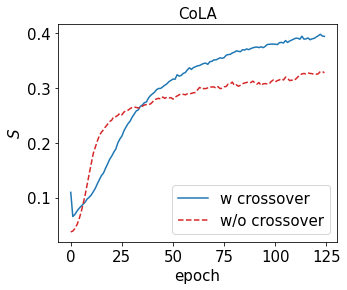}
      \caption{}
      \label{fig:cross_score_cola}
    \end{subfigure}%
    \begin{subfigure}{.24\textwidth}
      \centering
      \includegraphics[width=\linewidth]{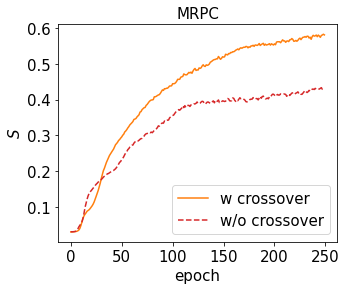}
      \caption{}
      \label{fig:cross_score_mrpc}
    \end{subfigure}%
    \hfill
    \begin{subfigure}{.24\textwidth}
      \centering
      \includegraphics[width=\linewidth]{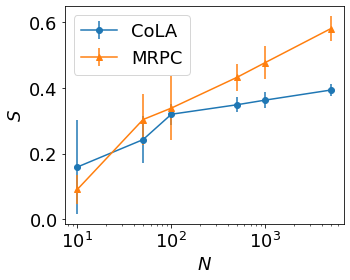}
      \caption{}
      \label{fig:N_score}
    \end{subfigure}%
    \begin{subfigure}{.24\textwidth}
      \centering
      \includegraphics[width=\linewidth]{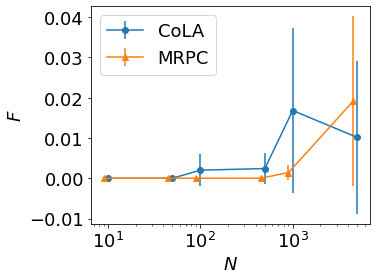}
      \caption{}
      \label{fig:N_nfool}
    \end{subfigure}
    \caption{Figure (a) and (b): Average score $S$ of generated sentences during epochs with and without crossover respectively for CoLA and MRPC task; Figure (c) and (d): Average final score $S$ and fraction of fooling sentences $F$ vs. number of generated sentences $N$ for Cola and MRPC tasks.}
\end{figure}

\subsection{Hyper-parameters analysis}

We describe here how boolean ($doCrossover$) and numerical ($N$ and $L$) hyper-parameters influence the performance of EFSG inspecting the mean score of sentences generated $S$ and the fraction of sentences fooling the model $F$. These quantities are computed at the last epoch $P$: for CoLA $P=125$ while for MRPC $P=250$. 

\subsubsection{Crossover contribute}

As described above, EFSG generates new sentences by modifying the earlier generation with \textit{mutation} and \textit{crossover} procedures. 
While \textit{mutations} are indispensable to generate always new sentences, \textit{crossover} can be avoided by picking for each new sentence only one \textit{Parent} with probability $S^{p-1}$ and performing mutations on it ($x_i^p \leftarrow Mutate(Parent_1, D_{task})$).
We evaluate here the contribution of crossover procedure by performing the same experiment twice, with and without crossover. 

Results prove that crossover is essential since EFSG generates sentences with an higher mean score $S$ than when EFSG is applied without crossover, on both CoLA and MRPC tasks. Figure~\ref{fig:cross_score_cola} and~\ref{fig:cross_score_mrpc} show how without crossover, $S$ reaches a plateau at about epoch $20$ for CoLA and about epoch $110$ for MRPC. 
In the following experiments we include crossover. 

\subsubsection{Number of sentences $N$ analysis}

The number of sentences $N$ generated at each epoch $p$ highly influences the performance of EFSG (setting $L=9$ for CoLA and $L=12$ for MRPC). 
The final average score $S$ increases as $N$ increase for both CoLA and MRPC (figure~\ref{fig:N_score}), but an high value of $N$ is essential to generate a relevant number of sentences fooling the model (figure~\ref{fig:N_nfool}). 
High values of standard deviation of $F$ are mainly due to a small number of experiments (five experiments each configuration). 

We set $N=5000$ for both CoLA and MRPC tasks for the following experiments.

\subsubsection{Length of sentences $L$ analysis}

The length of sentences $L$ is measured as the number of words in the sentence, counted splitting the string between spaces, while punctuation is previously removed. We check here how the value of $L$ influences the final average score $S$ and the fraction of fooling sentences $F$, fixing $N=5000$. 

\begin{figure}[t]
    \centering
    \begin{subfigure}{.24\textwidth}
      \centering
      \includegraphics[width=\linewidth]{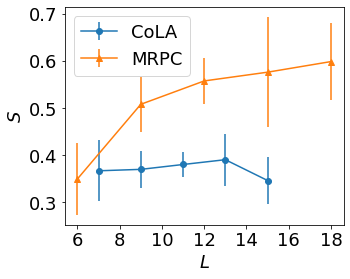}
      \caption{}
      \label{fig:L_score}
    \end{subfigure}
    \begin{subfigure}{.24\textwidth}
      \centering
      \includegraphics[width=\linewidth]{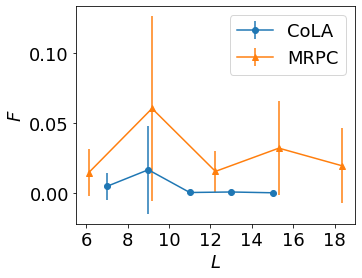}
      \caption{}
      \label{fig:L_nfool}
    \end{subfigure}
    \hfill
    \begin{subfigure}{.24\textwidth}
      \centering
      \includegraphics[width=\linewidth]{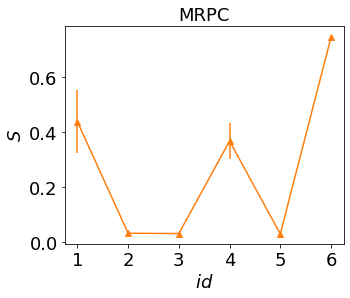}
      \caption{}
      \label{fig:id_score}
    \end{subfigure}%
    \begin{subfigure}{.24\textwidth}
      \centering
      \includegraphics[width=\linewidth]{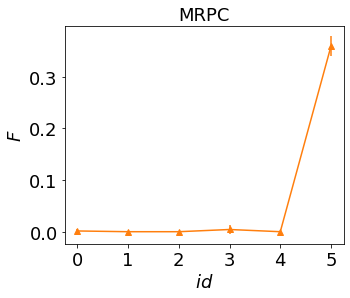}
      \caption{}
      \label{fig:id_nfool}
    \end{subfigure}
    \caption{Figure (a) and (b): Average final score $S$ and fraction of fooling sentences $F$ vs. length of generated sentences $L$ for Cola and MRPC tasks; Figure (c) and (d): Average final score $S$ and fraction of fooling sentences $F$ for different fixed sentences $\hat{x}$ for MRPC task.}
\end{figure}

For CoLA, generating sentences with different lengths $L$ is equally easy (figure~\ref{fig:L_score} and~\ref{fig:L_nfool}) since the values of $S$ and $F$ are almost independent of $L$. However, for MRPC, longer sentences reaches higher values of $S$, but this result is not reflected on $F$, that seems almost independent on $L$. 

We set $L=9$ (approximated average length of sentences in the original training set) for CoLA and $L=12$ (length of the fixed sentence $\hat{x}$) for MRPC for the following experiments.

\subsubsection{First sentence analysis}\label{sec:id}

When fooling a model on MRPC task, one sentence is fixed while the other is generated using EFSG. 
We investigate here how the performance of EFSG applied on tasks classifying couples of sentences is dependent on the fixed sentence $\hat{x}$. 

We select six different fixed sentences (Appendix B, table~\ref{tab:first_sen}, ids from $1$ to $6$) and we apply EFSG on MRPC task fixing one of them iteratively. For consistency reasons, we set $N=5000$ and $L$ equal to the length of the selected fixed sentence. Results (figure~\ref{fig:id_score} and~\ref{fig:id_nfool}) show how highly dependent the final result is on the fixed sentence. EFSG algorithm is not able to generate sentences that BERT base classifies as a paraphrase/semantic equivalence of some sentences (ids $2$, $3$ and $5$). Further investigations are required to understand if changing hyper-parameters or increasing the number of epochs $P$ is enough to fool the model, or if bigger improvements of EFSG are needed. 

\subsection{Model comparison}

To check if the generated fooling sentences are general weak spots or proper of a specific model, we investigate here how different models are fooled by the same sets of sentences, produced applying five times EFSG on BERT base fine-tuned (CoLA: $N=5000$ and $L=9$, MRPC: $N=5000$ and $L=12$). We report samples of sentences in Appendix C (table~\ref{tab:cola_n5000_l9} and~\ref{tab:mrpc_n5000_l12} respectively for CoLA and MRPC).
We test different fine-tunings of models, model sizes and model pre-trainings.

\subsubsection{Fine-tuning investigation}\label{sec:finet}

Due to the intrinsic randomness of the fine-tuning process, the same training set and hyper-parameters produce different fine-tuned models. 
We investigate if those models have the same weak spots (i.e. they are fooled by the same sentences).

Four different fine-tuned BERT base models are tested on the five generated sets previously described. 

\begin{figure}[t]
    \centering
    \begin{subfigure}{.24\textwidth}
      \centering
      \includegraphics[width=\linewidth]{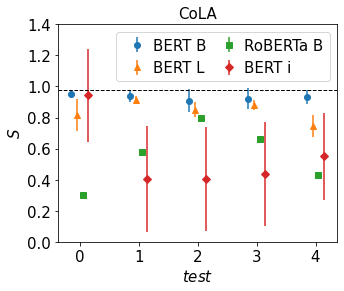}
      \caption{}
      \label{fig:comparison_score_cola}
    \end{subfigure}
    \begin{subfigure}{.24\textwidth}
      \centering
      \includegraphics[width=\linewidth]{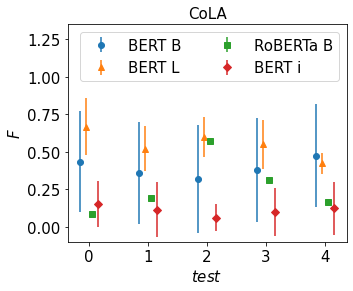}
      \caption{}
      \label{fig:comparison_M_cola}
    \end{subfigure}
    \hfill
    \begin{subfigure}{.24\textwidth}
      \centering
      \includegraphics[width=\linewidth]{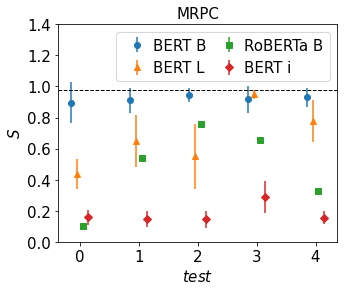}
      \caption{}
      \label{fig:comparison_score_mrpc}
    \end{subfigure}
    \begin{subfigure}{.24\textwidth}
      \centering
      \includegraphics[width=\linewidth]{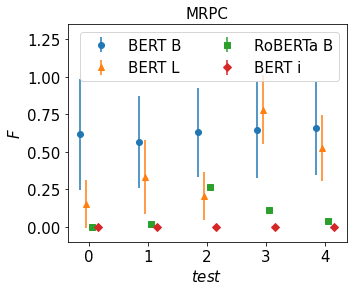}
      \caption{}
      \label{fig:comparison_M_mrpc}
    \end{subfigure}
    \caption{Figure (a) and (b): mean score $S$ and fraction of fooling sentences $F$ when LMs (BERT base, BERT large, RoBERTa and BERT improved) are tested on five sets of fooling sentences generated using EFSG on BERT base for CoLA and, Figure (c) and (d), for MRPC.}
\end{figure}

For CoLA, about $99\%$ of those sentences are classified wrongly by the four models. 
In detail (figure~\ref{fig:comparison_score_cola}, blue circles), the mean score of the five test sets is almost $1$ (the lower the better), and about $40\%$ of sentences generated fooling the first model fools also other models fine-tuned in the same way (figure~\ref{fig:comparison_M_cola}, blue circles). 

For MRPC $97\%$ of those sentences are classified wrongly by the four models.
Detailed results are similar to the ones obtained for CoLA (figure~\ref{fig:comparison_score_mrpc} and~\ref{fig:comparison_M_mrpc}, blue circles), with about $60\%$ of sentences fooling the four fine-tuned models. 

This result proves that the generated sentences are general and weakness of the models are independent of the fine-tuning process, for both CoLA and MRPC task. 

\subsubsection{Model size investigation: BERT large}

BERT has been released in multiple versions and sizes~\cite{turc2019wellread}, obtaining better and better results in many benchmark tasks. 
We evaluate here how the size of the model influences its robustness by testing five BERT large models fine-tuned using hyper-parameters described above in table~\ref{tab:hype} (evaluation accuracies already reported in figure~\ref{fig:eval_accs}) on the same five sets of sentences generated before. 
For CoLA, $84\%$ of sentences are misclassified, for MRPC, $77\%$ of them, less than BERT base. 
In detail (figure~\ref{fig:comparison_score_cola},~\ref{fig:comparison_M_cola},~\ref{fig:comparison_score_mrpc} and~\ref{fig:comparison_M_mrpc}, orange triangles) BERT large is better than BERT base everywhere but for CoLA, where an higher number of sentences fools BERT large for almost every test set.

This result shows that, even bigger BERT models are harder to fool than smaller ones, they still have many weak spots.

\subsubsection{Model pre-training investigation: RoBERTa}

RoBERTa differs from BERT mainly because of the pre-training procedure. 
We investigate here if this improvement, confirmed in figure~\ref{fig:eval_accs} by the evaluation accuracies, affects the robustness of the model to EFSG algorithm. We use the base version of RoBERTa.

For CoLA, RoBERTa misclassifies $48\%$ of sentences, for MRPC $46\%$ of them. In detail, mean scores and fraction of fooling sentences for both CoLA and MRPC tasks (figure~\ref{fig:comparison_score_cola},~\ref{fig:comparison_M_cola},~\ref{fig:comparison_score_mrpc} and~\ref{fig:comparison_M_mrpc}, green squares) indicate that RoBERTa base is not only more robust than BERT base, but also than BERT large in almost every experiment.

\subsection{Adversarial training}\label{sec:at}

Adversarial training is a standard defence strategy against adversarial attacks, where adversarial samples are used to train the model. 
Firstly, a model is fine-tuned and attacked. 
The successful samples (samples that fool the model) are collected and merged to the original training set. 
The original pre-trained model is then fine-tuned using the augmented training set and the obtained model is expected to be more resistant to the same kind of attacks. 

We firstly run EFSG on BERT base five times obtaining five sets of sentences by collecting every fooling sentence generated through the epochs for both tasks.  
We add these sets of adversarial samples to the original training sets and we use the new training sets to fine-tune five different models for each task. 
We call the new models \textit{BERT improved}. They exhibit no loss in accuracy with respect to BERT base, when tested on the original evaluation set (figure~\ref{fig:eval_accs}, red diamonds). 

To check if adversarial training is an effective defence strategy, we test the five models on the four sets used to adversarial train the four remaining models, obtaining misclassification of $46\%$ and $45\%$ of sentences respectively of CoLA and MRPC.
Detailed results (figure~\ref{fig:comparison_score_cola},~\ref{fig:comparison_M_cola},~\ref{fig:comparison_score_mrpc} and~\ref{fig:comparison_M_mrpc}, red diamonds) show how better are improved models with respect to the original ones.
The average scores $S$ and the fraction of fooling sentences $F$ of BERTs improved are lower than the other tested models for both CoLA and MRPC tasks. 
On MRPC, no sentences generated by using EFSG on BERT base fool BERT improved models. 

\paragraph{Fixed sentence $\hat{x}$ adversarial test}

Intuition suggests that, in the context of MRPC, an adversarial training set built adding as negative samples couples of sentences with the same fixed sentence $\hat{x}$ could result in a model learning, not how an adversarial sample looks like, but to classify negatively samples with that particular fixed sentence $\hat{x}$, regardless on the other one $x_i$. 
To check if this hypothesis is true, we test BERT improved with samples when $\hat{x}$ is $id=0$ in table~\ref{tab:first_sen}, on sets generated applying EFSG on BERT when the fixed sentences differs. 

Results (figure~\ref{fig:comparison_score_MRPC_id} and~\ref{fig:comparison_M_MRPC_id}, red diamonds) show that the performance of BERT improved is better (the lower the better) than the one obtained by the original BERT base (blue circles). However, the improvement is less significant than before, how we can notice comparing to figure~\ref{fig:comparison_score_mrpc} and~\ref{fig:comparison_M_mrpc} (when the test sets include samples with the same fixed sentence $\hat{x}$, where BERT improved is never fooled).

\subsection{EFSG applied on BERT large, RoBERTa base and BERT improved}

\begin{figure}[t]
    \centering
    \begin{subfigure}{.24\textwidth}
      \centering
      \includegraphics[width=\linewidth]{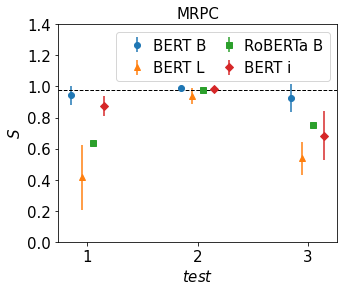}
      \caption{}
      \label{fig:comparison_score_MRPC_id}
    \end{subfigure}%
    \begin{subfigure}{.24\textwidth}
      \centering
      \includegraphics[width=\linewidth]{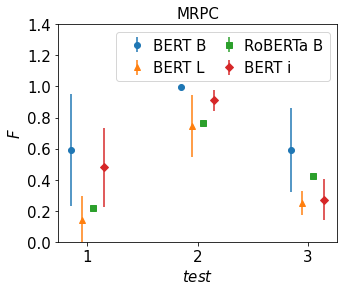}
      \caption{}
      \label{fig:comparison_M_MRPC_id}
    \end{subfigure}
    \hfill
    \begin{subfigure}{.24\textwidth}
      \centering
      \includegraphics[width=\linewidth]{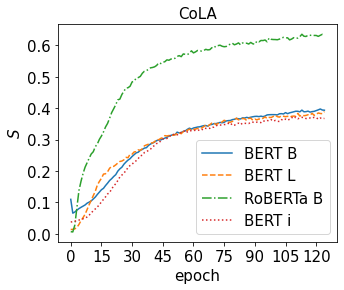}
      \caption{}
      \label{fig:pre_post_cola}
    \end{subfigure}%
    \begin{subfigure}{.24\textwidth}
      \centering
      \includegraphics[width=\linewidth]{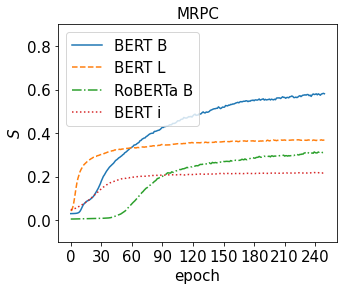}
      \caption{}
      \label{fig:pre_post_mrpc}
    \end{subfigure}
    \caption{Figure (a) and (b): mean score $S$ and fraction of fooling sentences $F$ when LMs (BERT base, BERT large, RoBERTa and BERT improved) are tested on three sets of sentences generated using EFSG on BERT base with different fixed sentences (MRPC task); Figure (c) and (d): mean score $S$ of generated sentences during the fooling process for BERT base, BERT large, RoBERTa base and BERT improved, for CoLA and MRPC respectively. }
\end{figure}

Finally, we run EFSG algorithm (with fixed hyper-paramenters) on BERT large, RoBERta base and BERT improved, comparing the results to BERT base.
Figure~\ref{fig:pre_post_cola} and~\ref{fig:pre_post_mrpc} show, respectively for CoLA and MRPC tasks, the mean score $S$ during the epochs. For CoLA, the mean score of BERT base, large and improved follow the same behaviours, suggesting an equal difficulty to fool the three models. RoBERTa, instead, gets higher scores. For MRPC, BERT improved is the stronger model, obtaining the lowest mean scores. 
Samples of generated fooling sentences are reported in tables in Appendix C (table~\ref{tab:cola_n5000_l9},~\ref{tab:mrpc_n5000_l12},~\ref{tab:cola_n5000_l9_large},~\ref{tab:cola_n5000_l9_rob} and~\ref{tab:cola_n5000_l9_imp}). No fooling sentence has been generated when EFSG is applied to BERT large, RoBERTa and BERT improved on MRPC task.

%% file: conclusions.tex
\section{Conclusion}\label{sec:conclusion}

We designed and tested EFSG, an evolutionary generator of adversarial fooling sentences. We test the robustness and generalization capabilities of some pre-trained LMs observing how they can resist to this kind of adversarial attack.
We tested different fine-tunings, model sizes and pre-training procedures, to understand if weak spots are general or dependent on one of those factors.
We used adversarial training as a defense procedure comparing the obtained improved models to the original ones. 

Due to the nature of EFSG, it can be applied to any language model and any binary classification task. 
We plan to apply the algorithm to every model implemented on Huggingface, for every dataset available. Future work will also involve the generalization of the algorithm to non-binary classification tasks (e.g. MNLI task~\cite{williams-etal-2018-broad}).

%% file: appendix.tex
\newpage
\section*{Appendix A: Best finetuning hyper-parameters}

\begin{table}[ht]
\centering
\begin{tabular}{cc|cccc}
Model      & Task & Learning rate    & Sequence length & Batch size & Epochs \\ \hline
BERT base  & CoLA & $10^{-5}$        & 64              & 64 & 3        \\
BERT large & CoLA & $10^{-5}$        & 64              & 12 & 3        \\
BERT base  & MRPC & $2\cdot 10^{-5}$ & 64              & 64 & 3        \\
BERT large & MRPC & $10^{-5}$        & 64              & 12 & 3       
\end{tabular}
\caption{Best finetuning hyperparameters}\label{tab:hype}
\end{table}

\section*{Appendix B: samples of fixed sentences $\hat{x}$ for MRPC}

\begin{table}[h]
\begin{adjustbox}{max width=\textwidth}
\begin{tabular}{c|l|c}
id & Fixed sencence $\hat{x}$                                                                                                                                                                & Length \\  \hline \hline
0  & The largest gains were seen in prices , new orders , inventories and exports & 12 \\ \hline
1  & The March downturn was the only break in what has been broad growth in services for & 20 \\
& the past 15 months                                                                         &      \\ \hline
2  & Other recommendations included a special counsel on oceans in the White House , creation & 26 \\ 
 & of regional ocean ecosystem councils and a national system to protect marine reserves &      \\ \hline
3  & The airline said the results showed that its recovery programme and other initiatives & 22 \\
& were helping to offset a continued deterioration in revenues                             &     \\  \hline
4  & The notification was first reported Friday by MSNBC                                                                                                                            & 8     \\ \hline
5  & The new law requires libraries and schools to use filters or lose federal financing & 14 \\ \hline
6  & China would become only the third nation to put a human in space & 13 \\ \hline
\end{tabular}
\end{adjustbox}
\caption{Fixed sentences $\hat{x}$ on MRPC}\label{tab:first_sen}
\end{table}

\section*{Appendix C: Samples of generated sentences}

\begin{table}[h]
\centering
\begin{tabular}{l|l}
Sentence & Score \\ \hline
resigned luscious students sang the phrase sick taste alex &0.989570 \\
round dog students loved the phrase over waffles arrived &0.986198 \\
the mailer students sang the phrase sylvia clowns the &0.984560
\end{tabular}
\caption{Sample of sentences generated by EFSG applied to BERT base on CoLA task ($N=5000$, $L=9$)}
\label{tab:cola_n5000_l9}
\end{table}

\begin{table}[h]
\centering
\begin{tabular}{l|l}
Sentence & Score \\ \hline
more results borrowed new wants I imported volume was seeing ordered ordered &0.979762 \\
more results arrive new been in imported the seen how I ordered &0.978105 \\
more results proved new america and imported meet in seeing for ordered &0.977136
\end{tabular}
\caption{Sample of sentences generated by EFSG applied to BERT base on MRPC task ($N=5000$, $L=12$)}
\label{tab:mrpc_n5000_l12}
\end{table}


\begin{table}[h]
\centering
\begin{tabular}{l|l}
Sentence & Score \\ \hline
oedipus yolk oedipus fatter sunny john john john propane &0.984796 \\
that semester cauliflo macaroni sandy t-shirts geordie nathan louis &0.982078 \\
siobhan brezhnev cauliflo macaroni caesar fish geordie good john &0.978200
\end{tabular}
\caption{Sample of sentences generated by EFSG applied to BERT large on CoLA task ($N=5000$, $L=9$)}
\label{tab:cola_n5000_l9_large}
\end{table}


\begin{table}[h]
\centering
\begin{tabular}{l|l}
Sentence & Score \\ \hline
satisfy please you'll the get will so must cops &0.999735 \\
play please all cooking savings ok so can the &0.999699 \\
propane please dress all please ok so do heidi &0.999659
\end{tabular}
\caption{Sample of sentences generated by EFSG applied to RoBERTa base on CoLA task ($N=5000$, $L=9$)}
\label{tab:cola_n5000_l9_rob}
\end{table}


\begin{table}[h]
\centering
\begin{tabular}{l|l}
Sentence & Score \\ \hline
those students this grant believed to liked froze the &0.984852 \\
all students the mailer believed holiday wanted left lost &0.980489 \\
those students there grant believed promised seconds froze amusing &0.979011
\end{tabular}
\caption{Sample of sentences generated by EFSG applied to BERT improved on CoLA task ($N=5000$, $L=9$)}
\label{tab:cola_n5000_l9_imp}
\end{table}
